\pdfoutput=1

\documentclass[11pt]{article}

\usepackage{acl}

\usepackage{times}
\usepackage{latexsym}
\usepackage{tikz}
\usepackage{amsmath}
\usepackage{graphicx}
\usepackage{float}
\usepackage{booktabs}
\usetikzlibrary{arrows, arrows.meta, positioning, calc}

\usepackage[T1]{fontenc}

\usepackage[utf8]{inputenc}

\usepackage{microtype}

\DeclareMathOperator*{\argmax}{arg\,max} 

\title{PK-ICR: Persona-Knowledge Interactive Multi-Context Retrieval for Grounded Dialogue}

\author{Minsik Oh* \And Joosung Lee \And Jiwei Li \And Guoyin Wang 
}

\begin{document}
\maketitle
\begingroup\def\thefootnote{*}\footnotetext{Please correspond to minsik@mskresearch.org.}\endgroup

\begin{abstract}
Identifying relevant persona or knowledge for conversational systems is critical to grounded dialogue response generation. However, each grounding has been mostly researched in isolation with more practical multi-context dialogue tasks introduced in recent works. We define Persona and Knowledge Dual Context Identification as the task to identify persona and knowledge jointly for a given dialogue, which could be of elevated importance in complex multi-context dialogue settings. We develop a novel grounding retrieval method that utilizes all contexts of dialogue simultaneously. Our method requires less computational power via utilizing neural QA retrieval models. We further introduce our novel null-positive rank test which measures ranking performance on semantically dissimilar samples (i.e. hard negatives) in relation to data augmentation.

\end{abstract}

\section{Introduction}

Effective conversation agents require external context as grounding information to enhance response generation. There has been much progress on each persona~\cite{majumder-etal-2020-like, persona-goal, persona-image, persona-response} and knowledge~\cite{yu-grounded-2022, dinan2018wizard, Zhao2020LowResourceKD, Liu2021ATL} grounded dialogue systems respectably. However, the combination of both and more unique contexts has not been studied, with limited interest in industry persona-based QA systems~\cite{hyundai-2017-patents, ericsson-2021-blog}.

\citet{feng-etal-2020-doc2dial, dinan2018wizard, moghe2018towards} have shown the importance of directly optimizing knowledge extraction in dialogue, while~\citet{zhang-etal-2018-personalizing, gu-etal-2021-persona-detecting,liu-etal-2020-impress} have shown the importance of directly optimizing for concrete persona. We further argue that in practical settings, it is more realistic to assume the utility of multiple contexts, with an explicit use-case being travel assistance agent~\cite{pkchat2022}.

\begin{figure}[t]
\centering
\colorbox{white}{\includegraphics[width=\columnwidth]{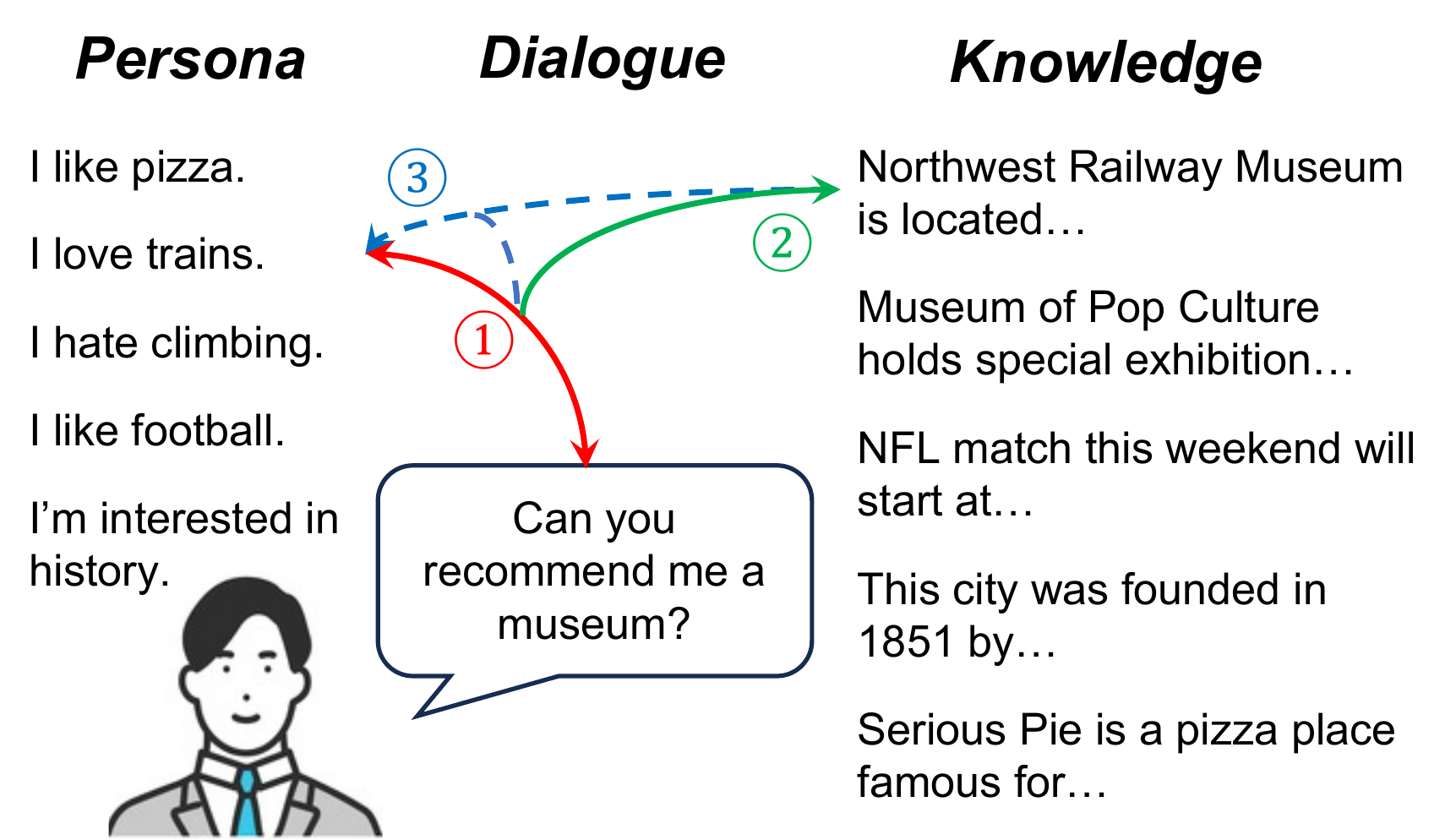}}
\caption{Illustration of dialogue component interactions regarding PK-ICR. In (1), dialogue is augmented with each persona to allow necessary interactions between persona and knowledge. (2) performs knowledge retrieval with (1). (3) performs precise persona scoring.}
\label{Interaction}
\end{figure}

Following the Knowledge Identification task in DIALKI~\cite{wu-etal-2021-dialki}, we define \textit{Persona and Knowledge Dual Context Identification} as the task to identify persona and knowledge jointly for a given dialogue. The task is similar to persona-based QA task in industry~\cite{hyundai-2017-patents, ericsson-2021-blog}, of creating a search engine based on persona, with exception of our study being in an interactive dialogue setting. We emphasize the specific interactions (Fig.~\ref{Interaction}) between persona, knowledge, and dialogue. We aim to formalize the nature of component-wise interactions via this research, resulting in enhanced multi-context retrieval methodology.

This separation of grounding retrieval tasks could be a particular benefit for multi-context dialogue, in which we can study complex context-wise interactions first, then apply the identified behavior as a sub-component of end-to-end systems. As a starting point, we re-purpose existing tasks and find that Question Answering (QA) is a good candidate~\cite{qa-2021}. This provides the benefit of reduced computation and streamlined architecture by reusing powerful retrieval models previously developed for diverse tasks.

We develop a framework that exploits this relation, of which an interesting aspect is combining persona and dialogue\footnote{Persona-augmented Dialogue} as a form of component augmentation. This may be of further utility in complex systems as each pertains to attributes and actions of the human respectively. Interestingly, our suggested augmentation method creates positive and hard negative samples which could be applied to enhance retrieval (Appendix~\ref{sec:back_app}). We introduce a novel evaluation methodology of the \textit{Null-positive Rank Test} (NRT) to quantify this trait. 

Our contributions are summarized as follows.\footnote{Code for our experiments is available : \url{https://github.com/minsik-ai/PK-ICR}}

1. \textbf{Persona and knowledge dual context retrieval methodology.} We enhance specific interactions between all components to successfully retrieve dialogue contexts. We achieve SOTA performance for both persona and knowledge retrieval.

2. \textbf{Framework for cross-task adaptation of dialogue context interactions.} We introduce a framework to benefit from existing performant retrieval models for complex dialogue grounding retrieval. Our zero-shot inference allows reduced computation (Table~\ref{tab:eff}) and streamlined architecture.

3. \textbf{Evaluating the hard-negative trait of Persona-augmented Dialogue.} We augment dialogue with persona to form an enhanced retrieval input, in which we observe hard negative traits. We introduce a novel test to isolate this trait, applicable to scenarios where semantically dissimilar samples are produced via data augmentation.

\section{Related Works}

"Knowledge-enhanced text generation"~\cite{zhu2022knowledge, yu2022survey} incorporates internal or external grounding contexts in tackling generative tasks such as dialogue or Q \& A. Our research significantly contributes to the development of sophisticated "knowledge selection" for external knowledge systems. Our work is the first to model how to effectively select multiple distinct types of grounding contexts (persona \& knowledge) for dialogue response generation.

To develop dialogue systems that rely on external knowledge information, open-domain dialogue datasets of Wizard of Wikipedia~\cite{dinan2018wizard} and PersonaChat~\cite{zhang2018personalizing} are most commonly studied. They consist of conversations that are grounded by Wikipedia knowledge and persona profile information, respectively. More recent datasets consist of conversations grounded by both persona \& knowledge~\cite{pkchat2022} or blended scenarios~\cite{shuster-etal-2020-dialogue, smith-etal-2020-put}. In line with the prior works, we treat persona and knowledge as different groundings with distinct characteristics and investigate semantic relations. 

Integrating either persona or knowledge bases with dialogue agents in isolation has been actively studied.~\citet{zhang-etal-2018-personalizing, majumder-etal-2020-like, persona-topical, persona-towards} for Persona, and~\citet{dinan2018wizard, Zhao2020LowResourceKD, Liu2021ATL, Li2020ZeroResourceKD, Ghazvininejad2017AKN} for knowledge. Persona-only method prohibits elaboration with detailed knowledge. In contrast, relevant knowledge might depend on the persona of the user. We address the limitations by studying all dialogue component interactions.

\textit{Knowledge Identification}~\cite{wu-etal-2021-dialki} task has been defined in recent papers stemming from knowledge-grounded dialogue. Our work aligns with the view in~\citet{wu-etal-2021-dialki} that context identification is a separately important task in an interactive dialogue setting, with similarities to open question answering~\cite{min-2019, chen-2017} and industry persona-based QA systems~\cite{hyundai-2017-patents, ericsson-2021-blog}. Our research expands upon the Knowledge Identification task to specify persona \& knowledge as dual contexts to be jointly retrieved from the dialogue.

\section{Methodology}

We maximize interactions between all components of a conversation turn for effective retrieval of dialogue groundings. Knowledge retrieval is a top-1 ranking task (Section~\ref{kr_subsection}), and persona retrieval is a point-wise scoring task with $1$ or $0$ true persona label (Section~\ref{pr_section}). We solve knowledge retrieval in a zero-shot manner, while we introduce \textit{null-positive rank test} to investigate the hard-negative traits of Persona-augmented Dialogue (Section~\ref{persona_finetune_tricks}).~\footnote{We note that each sequence - knowledge and persona retrieval - may be further optimized independently.}

\subsection{Knowledge Retrieval}
\label{kr_subsection}

We introduce a novel adaptation of dialogue components as QA prompts (example in Fig.~\ref{fig:QAform}). This form is selected to infer relations between all inputs of the grounded dialogue and to replicate short question and descriptive answer pairs.

\begin{equation}
E : \{Q_i, A_j\} = \{P_i + D, K_j\}
\label{QAFormEq}
\end{equation}

$E$ is input to our model. $Q_i, A_j, P_i, K_j$ are specific QA candidates and persona \& knowledge pairs. $D$ is the dialogue for the pairs. 

We then find the best knowledge for all pairs of $i$ and $j$ in a permutative manner and record the knowledge of the most aligned pair.

\begin{equation}
\textit{best}_i, \textit{best}_j = \argmax_{i \in {1...n}, j \in {1...m}} M_q\{P_i + D, K_j\}
\label{KtrueModel}
\end{equation}
\begin{equation}
\textit{true}_j = \textit{best}_j
\label{KSelection}
\end{equation}

$\textit{best}_i$, $\textit{best}_j$ are indices from best-scoring persona / knowledge pair. $\textit{true}_j$ is the index of predicted knowledge $K$. $\textit{best}_i$ is discarded.  $M_q$ is QA retrieval model for pair likelihood score. $n$ / $m$ is persona / knowledge count respectively.

\subsection{Persona Retrieval}
\label{pr_section}

Continuing from Section~\ref{kr_subsection}, we fine-tune the QA retrieval model using augmented persona and predicted true knowledge pairs only.\footnote{This results in additional reduced computation of $O(nm)$ to $O(n)$ for both training and inference. In effect, this decreases negative pairs from $3$M to $0.3$M with $10$x speedup.} We report that Persona-augmented Dialogue exhibits hard negative attributes (Section~\ref{persona_finetune_tricks}).

\begin{equation}
E' : \{Q_i, A_{\textit{true}}\} = \{P_i + D, K_{\textit{true}_j}\}
\label{KtrueInput}
\end{equation}

\begin{equation}
M_q \xrightarrow{E'_{\textit{train}}} M_f
\label{Finetune}
\end{equation}

$E'$ is input to our model similar to $E$, only difference being fixed true knowledge. $E'_{\textit{train}}$ is data from a separate training set formulated in the same manner as $E'$ with labeled true knowledge. $M_f$ is the fine-tuned model (Appendix~\ref{sec:exp_app}).

Next, we infer selected data pairs with $M_f$ to obtain the persona likelihood score. We avoid retrieving unrelated persona via a threshold.

\begin{equation}
p_i = M_f\{P_i + D, K_{true_j}\}
\label{Palign}
\end{equation}

\begin{equation}
\begin{aligned}
\textit{true}_i = \argmax_i
\begin{cases}
     p_i,& \text{if } p_i \geq p_{\textit{thres}} \\
    0,& \text{otherwise}
\end{cases}\\
\text{for }i \in {1...n}.
\end{aligned}
\label{PersonaEq}
\end{equation}

$p_i$ is the likelihood score for $P_i$. $p_{\textit{thres}}$ is the likelihood score threshold to remove persona that doesn't correspond to the dialogue turn. $\textit{true}_i$ is the index of the predicted true persona.

Finally, the retrieved grounding information is:

\begin{equation}
R : \{D, P, K\} = \{D, P_{\textit{true}_i}, K_{\textit{true}_j}\}
\label{Palign}
\end{equation}

$R$ is the retrieved true persona \& knowledge pair for the given dialogue turn.

\subsection{Null-positive Rank Test}
\label{persona_finetune_tricks}

\begin{figure}[t]
\centering
\includegraphics[width=0.6\columnwidth]{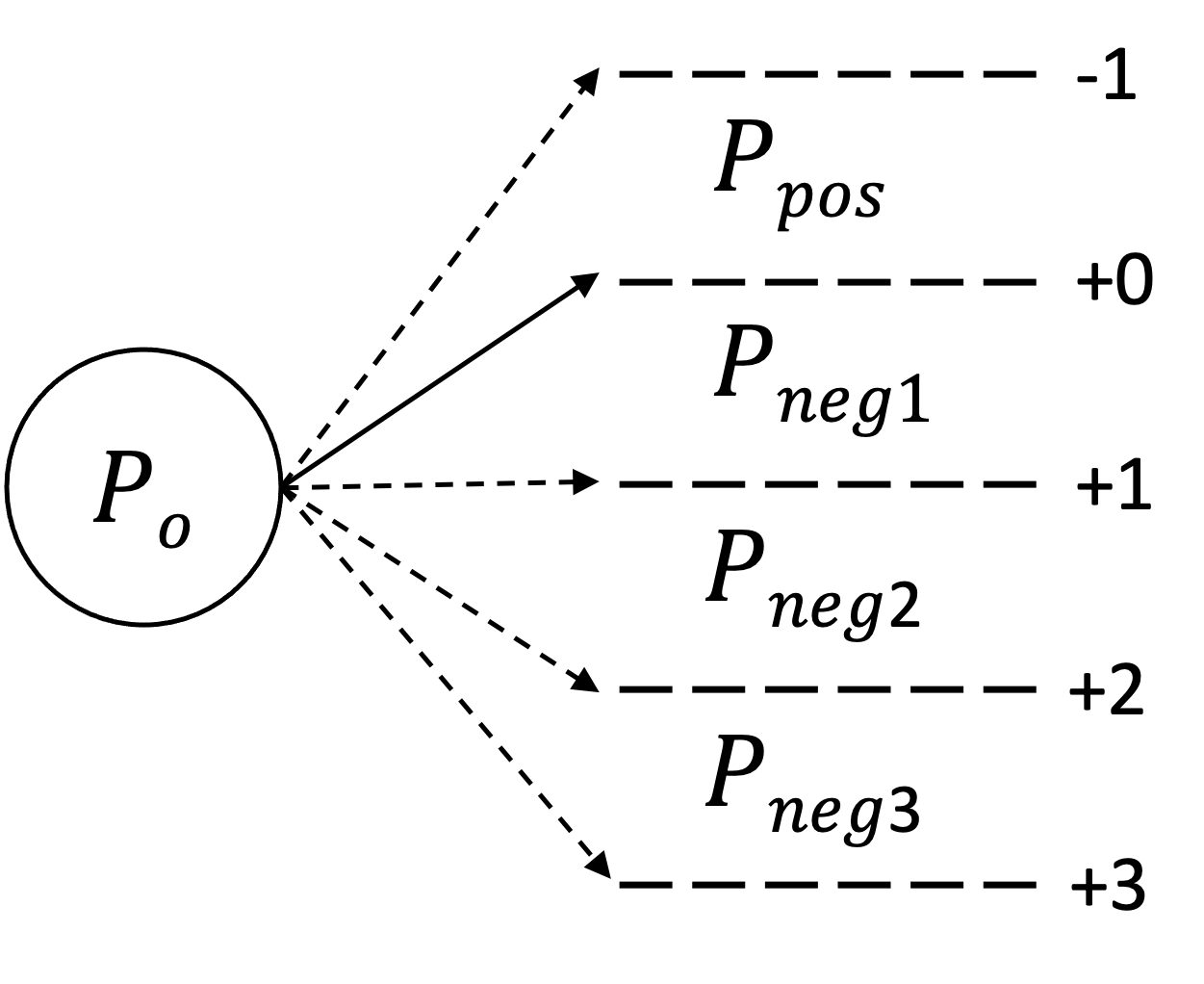}
\caption{Null-positive Rank Test (NRT). $P_o$, $P_{\textit{pos}}$, $P_{\textit{neg}_i}$ denote null-positive sample, positive and negative personas respectively. We omit augmentation $+ D$ in the figure for brevity. $r_{\textit{min}} = -1$ and $r_{\textit{max}} = +3$ in this figure. Arrows are possible positions for $P_o$. Numbers on the rightmost side are the null-positive adjusted rank values, being $0$ right below $P_{\textit{pos}}$ (example in Table~\ref{tab:nrt_samples}).}
\label{fig:nrt}
\end{figure}

We stress that fine-tuning model $M_q$ with Persona-augmented Dialogue ($P_i + D$) to create model $M_f$ is a specific choice. This is because the QA setup cannot be utilized without adjustments, due to the model scoring output skewing higher (Fig.~\ref{fig:persona_thres}). To analyze without inflated scores, we first interpret Persona-augmented Dialogue as hard negative sampling (Appendix~\ref{sec:back_app}), in which the augmentation produces non-trivially hard-to-distinguish samples.\footnote{Note that positive samples are also created (Table~\ref{tab:nrt_samples}). This augmentation is compatible with persona-only tasks.}

To evaluate the above observation, we present a novel methodology of \textbf{null-positive rank test} to quantify the inherent difficulty of ranking $P_i + D$ samples. Inspired by ranking metrics such as MRR, MAP, and NDCG, we utilize rank of a specific sample to compute model performance. This allows us to isolate the discriminative performance of the model corresponding to samples of interest, regardless of score output (Fig.~\ref{fig:nrt}, example in Table~\ref{tab:nrt_samples}).

We designate null-positive\footnote{"Null-positive" term corresponds to the fact that the ideal model should have no preference on how to score the likelihood of the null-positive sample, except that it should rank right below all positive sample(s). Another name considered was neutral rank test. In a real scenario, there could be multiple positive samples. Naturally, our metric weighs short rank distances less.} ($P_{o}$) sample as a baseline for the model. We measure the following: \textit{Can the model rank null-positive sample correctly in relation to non-trivially dissimilar augmented samples?} The "non-triviality" metric which computes the average distance of null-positive sample's rank from the ideal rank\footnote{Right below positive sample, above all negative samples.} is as follows:

\begin{equation}
\neg T = \frac{\sum_{r=r_{\textit{min}}}^{r_{\textit{max}}} n_r * \lvert r \rvert}{\sum_{r=r_{\textit{min}}}^{r_{\textit{max}}} n_r}
\label{eq:non-triv}
\end{equation}

Variants of the metric are in Appendix~\ref{sec:npr_var}. $\neg T$ is non-triviality metric, with lower values of the metric meaning the model ranks better. $n_r$ is the number of $P_o$ samples with adjusted rank $r$ (Fig.~\ref{fig:nrt}). We report "non-triviality" for each model $M_q$, $M_f$.

\section{Experiment Setup}
\label{sec:setup}

We utilize the Call For Customized Conversation~\cite{pkchat2022} dataset for which each conversation is built with both the user's persona and Wikipedia knowledge grounding information. We utilize multiple neural QA models trained on MS MARCO dataset~\cite{nguyen2016ms}. More details in Appendix~\ref{sec:exp_app}.

\section{Results}
\label{results}

\begin{table}[t]
\centering
    \begin{tabular}{cc}
        \toprule
        Model Type & Accuracy (\%) \\
        \midrule
        Baseline & 65.06 \\
        BERT-base & 11.78 \\
        Proto-gen~\cite{baseline-2022} & 85.18 \\
        $D$ \& $K_j$ & 79.26 \\
        $P_i$ \& $K_j$ & 84.62 \\
        $P_i + D$ \& $K_j$ & \textbf{94.69 (+29.63)} \\
        \bottomrule
    
    \end{tabular}
    \caption{Knowledge retrieval accuracy (cross-encoder) per asymmetric QA prompt. Zero-shot. Similar results for bi-encoder, which compares vectors (Table~\ref{knowledge_detail}).}
    \label{Ksummary}
\end{table}

\begin{table}[t]
    \centering
    \begin{tabular}{cc}
        \toprule
        Model Type & Accuracy (\%) \\
        \midrule
        Baseline & 86.86 \\
        BERT-base & 71.82 \\
        Proto-gen~\cite{baseline-2022} & 87.75 \\
        $D$ \& $P_i$ & 86.78 \\
        $P_i$ \& $K_{\textit{true}_j}$ & 86.75 \\
        $P_i + D$ \& $K_{\textit{true}_j}$ & 83.83 \\
        $P_i$ \& $K_{\textit{true}_j}$ (fine-tuned) & 89.12 \\
        $P_i + D$ \& $K_{\textit{true}_j}$ (fine-tuned) & \textbf{91.57 (+4.71)} \\
        \bottomrule
    
    \end{tabular}
    \caption{Persona retrieval accuracy (cross-encoder) per asymmetric QA prompt. Zero-shot unless fine-tuned. Similar results for bi-encoder (Table~\ref{persona_detail}).}
    \label{Psummary}
\end{table}

\begin{table}[t]
\centering
    \begin{tabular}{cccccc}
        \toprule
        Type & 0-Acc (\%) & $\neg T^2$ & $\neg T$ & $\neg T_{+}$ & $\neg T_{-}$\\
        \midrule
        Z.S. & 79.30 & 1.84 & 1.02 & 1.04 & 0.62 \\
        Ours & \textbf{86.81} & \textbf{1.66} & \textbf{0.97} & \textbf{0.96} & \textbf{0.56} \\
        \bottomrule
    \end{tabular}
    \caption{Null-positive rank test results for $P_i + D$ \& $K_{\textit{true}_j}$ cross-encoder models. Ours model is the fine-tuned variant, and Z.S. is Zero-Shot model. We report persona retrieval accuracy when $p_{thres} = 0$ (0-Acc) and variants of non-triviality (eq.~\ref{eq:non-triv},~\ref{eq:non-triv-sq},~\ref{eq:non-triv-pos},~\ref{eq:non-triv-neg}). Smaller non-triviality means superior ranking capability. Similar results for bi-encoder (Table~\ref{NPRresult-bi}).}
    \label{NPRresult}
\end{table}

\begin{figure}[t]
\centering
\includegraphics[width=.9\columnwidth]{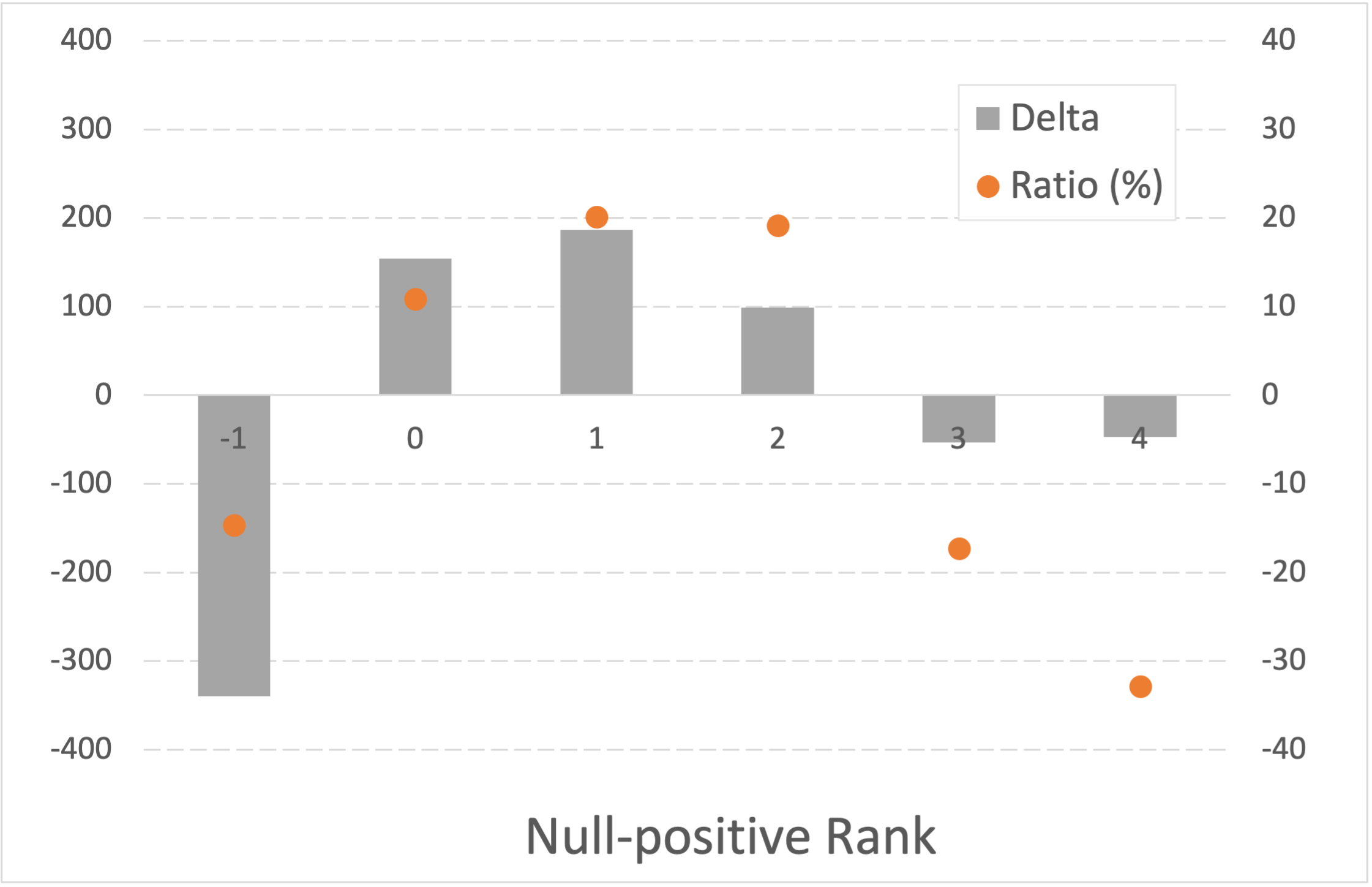}
\caption{Analysis of null-positive rank data for $P_i + D$ \& $K_{\textit{true}_j}$ cross-encoder model. Delta value is the change between Zero-shot model and Ours model in terms of sample count (left axis). Ratio value is delta value divided by sample count for Zero-shot, in \% (right axis).  We report movements of delta in correct directions for rank $-1$, $0$ and ranks with long distance $3$, $4$. Similar results for bi-encoder (Fig.~\ref{NullposCount-bi}).}
\label{NullposCount}
\end{figure}

\subsection{Knowledge Retrieval}
\label{kr_section}

We experiment with ablations of dialogue / persona / knowledge interactions and find permutative evaluation of eq.~\ref{QAFormEq} form yields best performance. Table~\ref{Ksummary} shows strong performance increase for our prompt input from dialogue-only model, confirming that all components of dialogue is important.\footnote{We verify that our cross-task adaptation of Q\&A is significantly stronger than NLI, STS, and NSP (Table~\ref{knowledge_detail}).}

\subsection{Persona Retrieval}

Table~\ref{Psummary} shows that fine-tuned $P_i + D$ model has the best performance. However, we observe low performance for the non-fine-tuned $P_i + D$ model. This is due to QA relationship of dialogue to true knowledge affecting the likelihood score (Fig.~\ref{fig:persona_thres}). Thus fine-tuning the model is a necessity to harness cross-domain adaptation.

\subsection{Null-positive Rank Test}

To verify our observation of the effectiveness of $P_i + D$, we perform null-positive rank test (Section~\ref{persona_finetune_tricks}). The performance of the model has increased in top-1 rank setting~($0$ threshold, 0-Acc)\footnote{This is performance on persona retrieval free from score-related effect (Fig.~\ref{fig:persona_thres}).} and all variants of non-triviality have improved for both models (Table~\ref{NPRresult},~\ref{NPRresult-bi}). We analyze sample count per rank (Fig.~\ref{NullposCount},~\ref{NullposCount-bi}).

\section{Discussions and Conclusion}

We introduce persona-knowledge dual context retrieval method PK-ICR in this paper. We perform QA-informed prompt-augmentations of data that successfully exploit the interactions between multiple dialogue components. We perform zero-shot top-1 knowledge retrieval and precise persona scoring. We present a novel evaluation method of null-positive rank test as to isolate the hard-negative effect of Persona-augmented Dialogue. We obtain SOTA results on both retrieval tasks of the Call For Customized Conversation benchmark and report the alignment of the non-triviality metric with threshold-free performance. With our research, we hope to stimulate readers to model dialogue context as an interactive whole of multiple components.

As the NLP community aims to tackle more complex dialogue systems, our methods may be further enhanced by sophisticated grounding contexts and interactions present in dialogue. Considering all components of dialogue, being persona, knowledge, and dialogue in a travel agent scenario, is crucial to obtain each grounding context required for accurate responses. We suggest two future directions for dialogue systems.

\begin{enumerate}
    \item One possible future direction is incorporating different forms of grounding such as persona/knowledge summaries, web searches, Wikipedia documents, and extracted sentences~\cite{wu-etal-2021-dialki} for multi-context interactions.
    \item Another would be extending our methodology to more sophisticated dialogue settings such as long-term memory~\cite{bae2022updated} or mutual persona~\cite{liu-etal-2020-impress}.
\end{enumerate}

Our persona-aware dialogue augmentation (\raisebox{.5pt}{\textcircled{\raisebox{-.9pt} {1}}} in Figure~\ref{Interaction}) is a form of modeling the human behavior in regards to Turing test~\cite{li2016personabased, vinyals2015neural}. We isolate effectiveness of our augmentation with successful Null-Positive Rank Test (Section~\ref{persona_finetune_tricks}). We suggest two future directions regarding persona-dialogue augmentation.

\begin{enumerate}
    \item One direction to pursue would be emphasizing different interactions, such as augmenting persona and knowledge first, or complex forms such as providing negative context in the augmentation themselves. Whether human-like behavioral modeling works better, or computationally tractable methods could be developed, is another interesting question.
    \item  Another direction to pursue would be advanced prompting, where specifying persona or knowledge via pre/post-fix prompts (i.e. "I'm thinking 'I like football'") provides more indicator information regarding each grounding and may improve performance.
\end{enumerate}

Finally, our novel Null-positive Rank Test (NRT, Section~\ref{persona_finetune_tricks}) is widely applicable to information retrieval and ranking models. We only trained the models with positive and negative dialogue augmentations, but interestingly, we report that the model improves ranking correctness for "neutral" (non-augmented) dialogue. This evaluation method could be directly used to compute whether the model obtains sophisticated ranking capabilities. This would be especially relevant in the context of data augmentation or other ranking tasks, including personalized search / QA. We recommend performing our test for future works.

\section{Limitations}

Our cross-task adaptation of dialogue grounding retrieval to QA task is limited in terms of the target task and our prompt construction. In addition, retrieval models informed by inductive bias for multi-context scenarios could further improve our methodology. 

We specifically study multi-context interactions and retrieval in dialogues, which is a relevant and novel problem for advancing broadly capable dialogue systems. As an extension to our research, future work could also report on modeling downstream generation tasks based on grounding interactions.

\section*{Acknowledgement}

We sincerely thank the ARR reviewers for their valuable insights.

\bibliography{anthology,custom}
\bibliographystyle{acl_natbib}

\newpage

\appendix
\onecolumn

\setcounter{table}{0}
\setcounter{figure}{0}
\renewcommand{\thetable}{A\arabic{table}}
\renewcommand{\thefigure}{A\arabic{figure}} 
\section{Samples}
\subsection{QA Cross-Task Adaptation Prompt Construction}
\label{sec:cross-domain}

\begin{figure}[H]
\centering
\begin{tikzpicture}
\node[draw, align=left] at (0,0) { Question : "\{I want to visit Seven Wonders of the\\ Ancient World.\} \{Wow, what is this?\}" \\ \\ Answer : "\{The Great Pyramid of Giza ... of the\\ Seven Wonders of the Ancient World, ...\}"};
\end{tikzpicture}
\caption{Resulting QA cross-task adaptation prompt of persona \& knowledge pair (eq.~\ref{QAFormEq}). Question form is "\{persona\} \{dialogue\}" while answer is "\{knowledge\}".}
\label{fig:QAform}
\end{figure}

\subsection{Persona-augmented Dialogue}
\label{sec:pad}

\begin{table}[H]
    \centering
    \begin{tabular}{p{.5\linewidth}cc}
        \toprule
        Persona-augmented Dialogue & Notation & Adj. Rank\\
        \midrule
        I like mountains, where to go for a hike? & $P_{\textit{pos}} + D$ & -1 \\
        \underline{where to go for a hike?} & $P_o = D$ & 0\\
        I like rock music, where to go for a hike? & $P_{\textit{neg1}} + D$ & +1\\
        I don't like pizza, where to go for a hike? & $P_{\textit{neg2}} + D$ & +2\\
        I don't like scary movies, where to go for a hike? & $P_{\textit{neg3}} + D$ & +3\\
        \bottomrule
    \end{tabular}
    \caption{We display ideal rank order for Persona-augmented dialogue ($P_i + D$) along with null-positive sample $P_o$ (underlined). The rank is adjusted to be $0$ for the ideal null-positive rank. This table corresponds to notations in Fig.~\ref{fig:nrt}.}
    \label{tab:nrt_samples}
\end{table}

\setcounter{table}{0}
\setcounter{figure}{0}
\renewcommand{\thetable}{B\arabic{table}}
\renewcommand{\thefigure}{B\arabic{figure}} 
\section{Null-Positive Rank Test Variants}
\label{sec:npr_var}

We introduce variants of non-triviality $\neg T$ metric (eq.~\ref{eq:non-triv}). Smaller numbers are better for all variants.

\begin{itemize}

\item $\neg T_+$ to only observe positive rank displacements.

\begin{equation}
\neg T_+ = \frac{\sum_{r=0}^{r_{\textit{max}}} n_r * \lvert r \rvert}{\sum_{r=0}^{r_{\textit{max}}} n_r}
\label{eq:non-triv-pos}
\end{equation}

\item $\neg T_-$ to only observe negative rank displacements.

\begin{equation}
\neg T_- = \frac{\sum_{r=r_{\textit{min}}}^{0} n_r * \lvert r \rvert}{\sum_{r=r_{\textit{min}}}^{0} n_r}
\label{eq:non-triv-neg}
\end{equation}

\item $\neg T^2$ similar to how Mean Squared Error relates to Mean Absolute Error.
\begin{equation}
\neg T^2 = \frac{\sum_{r=r_{\textit{min}}}^{r_{\textit{max}}} n_r * r^2}{\sum_{r=r_{\textit{min}}}^{r_{\textit{max}}} n_r}
\label{eq:non-triv-sq}
\end{equation}

\item $\neg T_{\textit{weighted}}$ to provide constant weights for each rank.

\begin{equation}
\neg T_{\textit{weighted}} = \frac{\sum_{r=r_{\textit{min}}}^{r_{\textit{max}}} w_r * n_r * \lvert r \rvert}{\sum_{r=r_{\textit{min}}}^{r_{\textit{max}}} w_r * n_r}
\label{eq:non-triv-w}
\end{equation}

\end{itemize}

\setcounter{table}{0}
\setcounter{figure}{0}
\renewcommand{\thetable}{C\arabic{table}}
\renewcommand{\thefigure}{C\arabic{figure}} 

\section{Computational Efficiency}
\label{sec:eff_app}

\citet{baseline-2022} utilizes 2 BART models for input and persona \& knowledge groundings.\footnote{2 BART encoders and 1 decoder for generation (may positively affect retrieval).} In contrast, PK-ICR utilizes 1 MiniLM~\cite{wang2020minilm} cross-encoder retrieval model for computing similarity scores.~\citet{baseline-2022} concatenates all groundings as one input, while PK-ICR groups them into pairs. Our zero-shot knowledge retrieval (Section~\ref{kr_subsection}) doesn't require any training, while~\citet{baseline-2022} trains for a maximum of 15 epochs.

\begin{table}[H]
\centering
    \begin{tabular}{ccc}
        \toprule
        Metrics & \citet{baseline-2022} & Ours \\
        \midrule
        Train Samples & $1.1$M & $633$K (Section~\ref{pr_section}) \\
        Model Params & $210$M & $33$M \\
        \bottomrule
    \end{tabular}
    \caption{Computational effort required for the methods.}
    \label{tab:eff}
\end{table}

\section{Experiment Setup}
\label{sec:exp_app}

We utilize Call For Customized Conversation \cite{pkchat2022} dataset for evaluation and fine-tuning, which has 10 knowledge candidates and 5 persona candidates per dialogue. We utilize 12 layer MiniLM~\cite{wang2020minilm} (33M params) cross-encoder trained on MS MARCO\footnote{MRR@10 on MS MARCO Dev Set: 39.02}~\cite{nguyen2016ms} from Sentence-BERT library~\cite{sbert2019} and DistillBERT (66M params) TAS-B~\cite{Hofstaetter2021_tasb_dense_retrieval} bi-encoder model trained on same data\footnote{MRR@10 on MS MARCO Dev Set: 34.43}. This data is for semantic search, with trained models evaluating short questions and long passages together. In addition, we report baseline performance on DPR~\cite{karpukhin2020dense} models\footnote{Bi-encoder model with separate question and answer encoders.} (110M params) trained on NQ~\cite{kwiatkowski-etal-2019-natural} dataset\footnote{NQ test set Accuracy@20: 78.4, Accuracy@100: 85.4}, with a dummy short segment of "Title", and treating Knowledge as long answer segment. For persona search (eq.~\ref{Finetune},~\ref{PersonaEq}), we fine-tune for 2 epochs, 32 batch size, and sigmoid activation function with Binary Cross Entropy (cross-encoder) / Cosine Similarity (bi-encoder) Loss with $p_{\textit{thres}} = 0.5$. We list the official evaluation results on the test data. For MobileBERT (25M params) and BERT-base (110M params), we evaluate with Next Sentence Prediction task. We experiment with DistillRoBERTa (82M params) STS\footnote{STSbenchmark test performance: 87.92} and NLI\footnote{Accuracy on MNLI mismatched set: 83.98, we compare with \textit{entailment} score.} cross-encoder models. We work with RTX 3090 NVIDIA GPU.

\setcounter{table}{0}
\setcounter{figure}{0}
\renewcommand{\thetable}{E\arabic{table}}
\renewcommand{\thefigure}{E\arabic{figure}}

\section{Persona Retrieval Threshold Experiments}
\label{sec:persona_thres}

\begin{figure}[H]
    \centering
    \includegraphics[width=.6\columnwidth]{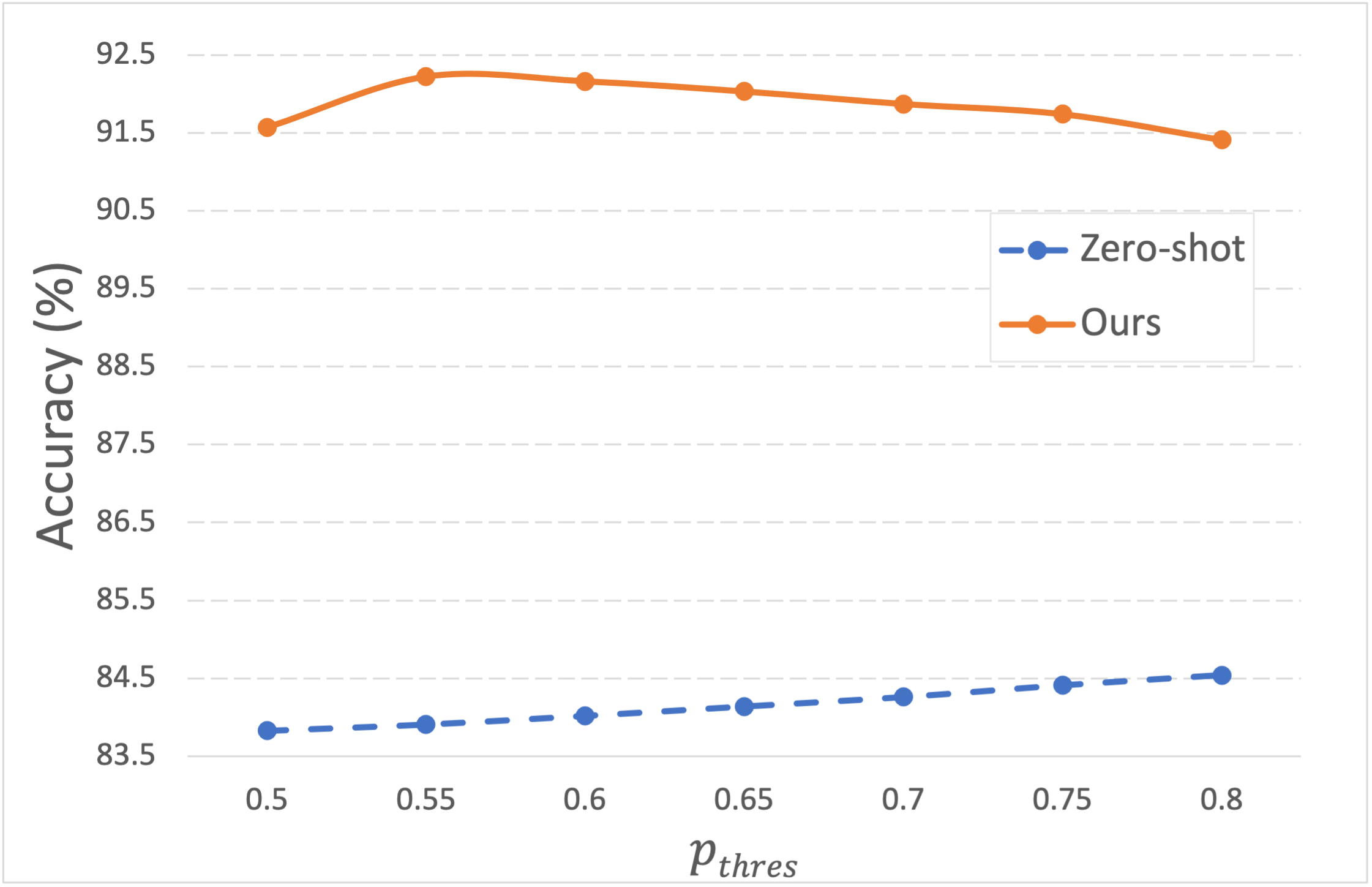}
    \caption{Persona threshold ablation experiments with $P_i$ \& $K_{\textit{true}_j}$ cross-encoder model. We report persona accuracy. $p_{\textit{thres}}$ is defined in eq.~\ref{PersonaEq}. Dotted line correspond to Zero-shot model, and solid line is our best model. We find visible peak at $0.55$ with our best model while Zero-shot model performance keeps increasing $> 0.8$.}
    \label{fig:persona_thres}
\end{figure}

\setcounter{table}{0}
\setcounter{figure}{0}
\renewcommand{\thetable}{F\arabic{table}}
\renewcommand{\thefigure}{F\arabic{figure}}  

\section{Null-Positive Rank Test for Bi-Encoder}

\begin{table}[H]
\centering
    \begin{tabular}{cccccc}
        \toprule
        Type & 0-Acc (\%) & $\neg T^2$ & $\neg T$ & $\neg T_{+}$ & $\neg T_{-}$\\
        \midrule
        Z.S. & 77.90 & 4.27 & 1.59 & 1.79 & 0.55 \\
        Ours & \textbf{85.53} & \textbf{1.95} & \textbf{0.99} & \textbf{0.98} & \textbf{0.50} \\
        \bottomrule
    \end{tabular}
    \caption{Null-positive rank test results for $P_i + D$ \& $K_{\textit{true}_j}$ bi-encoder models. Ours model is the fine-tuned variant, and Z.S. is Zero-Shot model. We report persona retrieval accuracy when $p_{\textit{thres}} = 0$ (0-Acc) and variants of non-triviality (eq.~\ref{eq:non-triv},~\ref{eq:non-triv-sq},~\ref{eq:non-triv-pos},~\ref{eq:non-triv-neg}). Smaller non-triviality means superior ranking capability.}
    \label{NPRresult-bi}
\end{table}

\begin{figure}[H]
\centering
\includegraphics[width=.6\columnwidth]{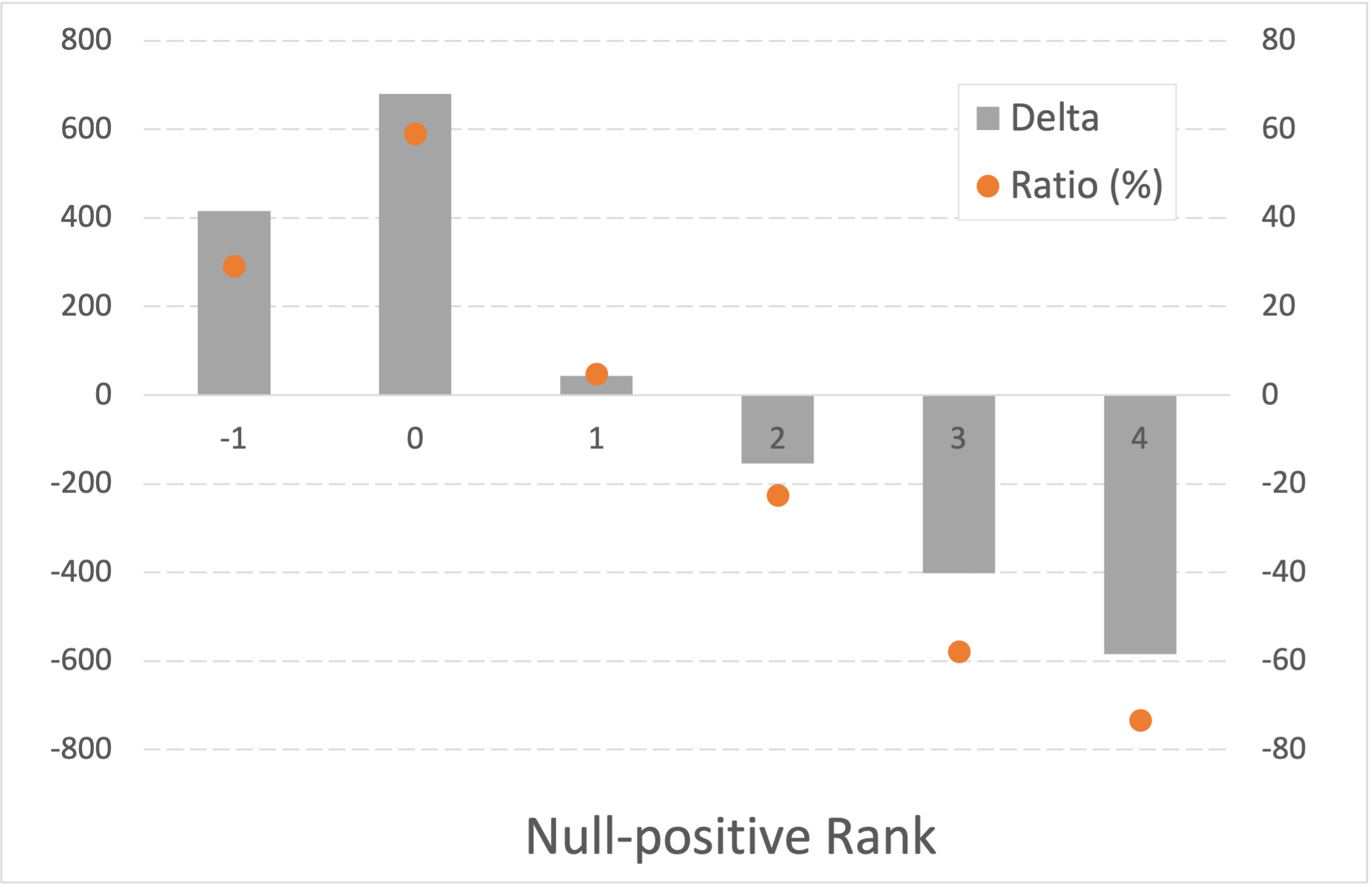}
\caption{Analysis of null-positive rank data for $P_i + D$ \& $K_{\textit{true}_j}$ bi-encoder model. Delta value is the change between Zero-shot model and Ours model in terms of sample count (left axis). Ratio value is delta value divided by sample count for Zero-shot, in \% (right axis). We report large movements of delta in correct directions for rank $0$ and ranks with long distance $3$, $4$.}
\label{NullposCount-bi}
\end{figure}
\setcounter{table}{0}
\setcounter{figure}{0}
\renewcommand{\thetable}{I\arabic{table}}
\renewcommand{\thefigure}{I\arabic{figure}} 
\section{Background: Ranking and Hard Negative Sampling}
\label{sec:back_app}

Text ranking is a task to generate an ordered list of texts in response to a query~\cite{lin2021ranking}. It is a core task of information retrieval (IR) where you obtain a ranked list of samples ordered by estimated relevance to the query. We introduce widely accepted neural approaches, 'cross-encoder' and 'bi-encoder'. We will also describe 'hard negative sampling', a data-centric approach to improve retrieval models.

For cross-encoder~\cite{nogueira2019mono}, query and a single sample are concatenated by '[SEP]' token as an input to the model, resulting in a relevance score (FFN output of the '[CLS]' token representation) for the specific sample. We note that this setup is similar to sentence-wise classification settings presented in~\citet{devlin2018bert}. Then, the samples are ordered by relevance score to produce the final ranked list.

For bi-encoder~\cite{sbert2019}, we generate dense vectors (sentence  embeddings) per each query and each sample. This is obtained via '[CLS]' token representation of a specially fine-tuned model\footnote{Supervised learning via Siamese Network.}, with a single query or sample input. The representations are then compared as pairs via cosine-similarity or dot-product to compute relevance scores. While original bi-encoder setup computes sentence-wise similarity (STS), we utilize models fine-tuned on QA data (Appendix~\ref{sec:exp_app}).

Hard negative sampling (also known as hard negative mining) is a technique to obtain specific samples (hard negatives) that are difficult to distinguish from positive samples, yet have a different label. The hard negative samples are then incorporated during model fine-tuning to improve model capabilities. For example, in the context of ranking, non-relevant texts scoring high by how many keywords match~\cite{xiong2020approximate} may be considered hard negatives.~\citet{xiong2020approximate,luan2021sparse,lin-etal-2021-batch} have demonstrated that hard negative sampling improves ranking models considerably.

\setcounter{table}{0}
\setcounter{figure}{0}
\renewcommand{\thetable}{H\arabic{table}}
\renewcommand{\thefigure}{H\arabic{figure}}  
\section{Detailed Results}
\label{sec:detailed_results}

More experiments are listed here. Our bi-encoder and cross-task experiments confirm our findings in Section~\ref{results}. Explanation of the models in Appendix~\ref{sec:back_app}.

\begin{table}[H]
\centering
    \begin{tabular}{cc}
        \toprule
        Model Type & Accuracy (\%) \\
        \midrule
        Baseline & 65.06 \\
        MobileBERT & 9.49 \\
        BERT-base & 11.78 \\
        Proto-gen~\cite{baseline-2022} & 85.18 \\
        \midrule
        NLI (cross-encoder) & 17.96 \\
        STS (cross-encoder) & 51.33 \\
        $D$ \& $K_j$ (cross-encoder) & 79.26 \\
        \midrule
        $P_i$ \& $K_j$ (DPR) & 75.54 \\
        $P_i$ \& $K_j$ (bi-encoder) & 80.73 \\
        $P_i$ \& $K_j$ (cross-encoder) & 84.62 \\
        $P_i + D$ \& $K_j$ (DPR) & 83.98 \\
        $P_i + D$ \& $K_j$ (bi-encoder) & 92.67 \\
        $P_i + D$ \& $K_j$ (cross-encoder) & \textbf{94.69} \\
        \bottomrule
    
    \end{tabular}
    \caption{Knowledge retrieval results, all models are zero-shot. We report top-1 knowledge retrieval accuracy per asymmetric QA prompt. D, K, P each refer to dialogue, knowledge and persona.}
    \label{knowledge_detail}
\end{table}

\begin{table}[H]
    \centering
    \begin{tabular}{cc}
        \toprule
        Model Type & Accuracy (\%) \\
        \midrule
        Baseline & 86.86 \\
        MobileBERT & 86.86\\
        BERT-base & 71.82 \\
        Proto-gen~\cite{baseline-2022} & 87.75 \\
        $D$ \& $P_i$ (cross-encoder) & 86.78 \\
        \midrule
        $P_i$ \& $K_{\textit{true}_j}$ (DPR) & 75.54 \\
        $P_i$ \& $K_{\textit{true}_j}$ (bi-encoder) & 78.64 \\
        $P_i$ \& $K_{\textit{true}_j}$ (cross-encoder) & 86.75 \\
        $P_i$ \& $K_{\textit{true}_j}$ (cross-encoder, fine-tuned) & 89.12 \\
        \midrule
        $P_i + D$ \& $K_{\textit{true}_j}$ (DPR) & 74.76 \\
        $P_i + D$ \& $K_{\textit{true}_j}$ (bi-encoder) & 77.90 \\
        $P_i + D$ \& $K_{\textit{true}_j}$ (cross-encoder) & 83.83 \\
        $P_i + D$ \& $K_{\textit{true}_j}$ (bi-encoder, fine-tuned) & 85.55 \\
        $P_i + D$ \& $K_{\textit{true}_j}$ (cross-encoder, fine-tuned) & \textbf{91.57} \\
        \bottomrule
    
    \end{tabular}
    \caption{Persona retrieval results, models are zero-shot unless fine-tuned. We report persona retrieval accuracy per asymmetric QA prompt. We do not fine-tune DPR model due to implementation limitations. D, K, P each refer to dialogue, knowledge and persona.}
    \label{persona_detail}
\end{table}

\newpage

\setcounter{table}{0}
\setcounter{figure}{0}
\renewcommand{\thetable}{I\arabic{table}}
\renewcommand{\thefigure}{I\arabic{figure}} 
\section{Retrieval Output Samples}
\label{sec:outputs}

We list some of the retrieved outputs with our best model in Table~\ref{tab:result_samples}.

\begin{table*}[h]
    \centering
    \begin{tabular}{p{.2\linewidth}p{.2\linewidth}p{.5\linewidth}}
        \toprule
        dialogue $D$ & persona $P_{\textit{true}}$ & knowledge $K_{\textit{true}}$\\
        \midrule
        I think I've been there before but I don't remember the name of this place. & I am fond of Modernist architechure. & The Casa de les Punxes or Casa Terradas is a building designed by the Modernista architect Josep Puig I Cadafalch. Located in the intersection between the streets of Rosselló, Bruc and the Avinguda Diagonal in the Barcelona Eixample area.\\
        \midrule
        How much this railway line costed in those times? & I love railway. & Because of the difficult physical conditions of the route and state of technology, the construction was renowned as an international engineering achievement, one that cost US\$8 million and the lives of an estimated 5,000 to 10,000 workers. \\
        \midrule
        Who built this rail line? & I love railway. & The line was built by the United States and the principal incentive was the vast increase in passenger and freight traffic from eastern USA to California following the 1849 California Gold Rush.\\
        \midrule
        What's the highest point in the Mulanje Massif? & I like to climbing up the elevations on my neighborhood to take a look around. & Sapitwa Peak, the highest point on the massif at 3,002 m, is the highest point in Malawi.\\
        \midrule
        Who was the first explorer to find this mountain? & I have fantasies of being a Livingstone type explorer. & The first European to report seeing the Massif was David Livingstone in 1859, but archeological investigation reveals evidence of human visits to the Massif from the Stone Age onwards. \\
        \midrule
        Now I remember, can you tell me some characteristics of this channel? & N / A & And may be the oldest canal in England that is still in use. It is usually thought to have been built around AD 120 by the Romans, but there is no consensus among authors. \\
        \bottomrule
    \end{tabular}
    \caption{persona, knowledge and dialogue retrieved examples from our best model.}
    \label{tab:result_samples}
\end{table*}

\end{document}